\documentclass[final,5p,times,twocolumn,authoryear]{elsarticle}

\usepackage{cite}

\usepackage{graphicx}
\DeclareGraphicsExtensions{.pdf,.jpeg,.png}
\usepackage{booktabs}
\usepackage[hypcap]{caption}
\usepackage{float}
\usepackage{url}
\usepackage{hyperref}
\usepackage{color}
\newcommand{\fref}[1]{Fig.~\ref{#1}}
\newcommand{\tref}[1]{Table ~\ref{#1}}
\newcommand{\eref}[1]{(\ref{#1})}
\usepackage[
    type={CC},
    modifier={by-nc-nd},
    version={4.0},
    imagewidth=3em
]{doclicense}

\usepackage[cmex10]{amsmath}
\makeatletter
\newcommand{\mypm}{\mathbin{\mathpalette\@mypm\relax}}
\newcommand{\@mypm}[2]{\ooalign{%
  \raisebox{.5\height}{$#1+$}\cr
  \smash{\raisebox{-.3\height}{$#1-$}}\cr}}
\makeatother
\newcommand{\bpm}{\sbox0{$1$}\sbox2{$\scriptstyle\mypm$}
  \raise\dimexpr(\ht0-\ht2)/2\relax\box2 }

\usepackage{array}

\usepackage{microtype}
\DisableLigatures{encoding = *, family = * }

\journal{Computer Speech and Language}

\begin{document}

\begin{frontmatter}



\title{On the Relevance of Auditory-Based Gabor Features for Deep Learning in \\ Robust Speech Recognition}

\author[UniOL,H4A]{Angel Mario Castro Martinez\corref{cor1}} 
\ead{angel.castro@uni-oldenburg.de}

\author[JHU]{Sri Harish Mallidi}
\ead{mallidi@jhu.edu}

\author[JHU]{Bernd T. Meyer}
\ead{bernd.t.meyer@jhu.edu}

\address[UniOL]{Department f\"ur medizinische Physik und Akustik, Carl von Ossietzky Universit\"at Oldenburg, Germany}
\address[H4A]{Exzellenzcluster Hearing4all, Germany}

\address[JHU]{Center for Language and Speech Processing, Johns Hopkins University, Baltimore, MD, USA}

\fntext[cor1]{Corresponding author}

\tnotetext[]{\doclicenseImage  \doclicenseText}

\begin{abstract}

Previous studies support the idea of merging auditory-based Gabor features with deep learning architectures to achieve robust automatic speech recognition, however, the cause behind the gain of such combination is still unknown. We believe these representations provide the deep learning decoder with more discriminable cues. Our aim with this paper is to validate this hypothesis by performing experiments with three different recognition tasks (Aurora\,4, CHiME\,2 and CHiME\,3) and assess the discriminability of the information encoded by Gabor filterbank features. Additionally, to identify the contribution of low, medium and high temporal modulation frequencies subsets of the Gabor filterbank were used as features (dubbed LTM, MTM and HTM respectively). With temporal modulation frequencies between $16$ and $25$\,Hz, HTM consistently outperformed the remaining ones in every condition, highlighting the robustness of these representations against channel distortions, low signal-to-noise ratios and acoustically challenging \textit{real-life} scenarios with relative improvements from $11$ to $56$\% against a Mel-filterbank-DNN baseline. To explain the results, a measure of similarity between phoneme classes from DNN activations is proposed and linked to their acoustic properties. We find this measure to be consistent with the observed error rates and highlight specific differences on phoneme level to pinpoint the benefit of the proposed features. 

\end{abstract}


\end{frontmatter}

%\thanks{Manuscript received December 1, 2015; revised September 17, 2014. Corresponding author: A. Castro Martinez (email: angel.castro@uni-oldenburg.de).}}

\section{Introduction}
\label{intro}

Over the last decade there have been major advances in automatic speech recognition (ASR), which mainly have promoted ubiquitous speech enhanced technologies in our daily lives. 

The approaches to transcribe speech into words have changed in several ways throughout the years. Not even 10 years ago, the use of hidden Markov models (HMM) to represent speech as sequence of time-varying states and Gaussian mixture models (GMM) to statistically fit the acoustic input to these states was widely adopted as the standard for ASR; actually, due to the implementation of discriminative training methods to optimize HMM classification \citep{Povey2005}, \citep{Xiadong2008}, GMM-HMM recognizers yielded the best performance among other systems.

Recently, however, deep neural networks (DNNs) have successfully replaced GMMs in both small and large vocabulary tasks \citep{Mohamed2011,Mohamed2012},\citep{Pan2012}, \citep{Seide2011}, \citep{Sainath2011} for reasons better explained in \citep{Hinton2012}.

In spite of all aforementioned advances, ASR performance still lags far behind its human counterpart, especially in noisy and reverberant environments, thereby preventing the further development of technologies empowered by ASR, regardless how appealing or necessary they might be. To bridge this gap, researchers have focused on two different, but not mutually exclusive, strategies:  developing better back-ends and extracting more informative discriminable features. For a thorough overview of noise-robust techniques successfully implemented in ASR research, refer to \citep{Li2014}.

Concerning DNNs, one recipe to accomplish the former goal is relatively straightforward, it involves the trade-off between lowering the error and generalization (much like many other machine learning algorithms) and depends on how the system performs on a cross-validation set. On the one hand if the purpose is to minimize the loss function, the course of action is to increase the model complexity, i.e. increase the number of parameters either the number of neurons per layer or the depth of the network by adding additional layers. 

On the other hand if the loss function of the cross-validation set increases (a situation known as overfitting), there is a need for more training examples. In some cases the amount of available data is limited and despite the vast advances in computing software and hardware, training times do not scale well on deep architectures; for those cases, the second objective seems to be more accessible. As the healthy human ear is still unmatched in its robustness \citep{Lippmann1997}, \citep{Scharenborg2007}, \citep{Meyer2010}, mimicking its principles improves existing feature extraction methods for ASR; better representations, in turn, could potentially lead to a broader understanding of the underlying principles of human auditory processing.

The use of feature extraction techniques inspired by the auditory system have previously demonstrated a boost in speech recognition performance. Even the most widely used Mel frequency cepstral coefficients (MFCC) or features resulting from the perceptive linear predictive (PLP) analysis of speech \citep{Hermansky1990}, intrinsically implement biological findings. Owing to the glottal source of speech low frequencies have more energy therefore a pre-emphasis stage equalizes the signal power; both features use a different scale for frequency warping derived from psychoacoustic measurements (the Mel scale for MFCC and the Bark scale for PLP). Non-linear functions are applied for amplitude compression mimicking loudness perception of the auditory system (the logarithm for MFCC and an intensity-loudness power law for PLP features, respectively). Additionally, in PLP several properties of hearing concerning asymmetries in frequency selectivity and equal loudness are simulated in more detail resulting into a closer auditory-like spectrogram than the log-Mel used in MFCC.

In order to increase the recognizer robustness to channel distortions and other convolutional noise sources, MFCC and PLP features were extended by processing mechanisms such as cepstral mean normalization and RASTA processing \citep{Hermansky1994}, the latter consists of bandpass filtering the compressed spectral amplitudes to emphasize transients, imitating the auditory periphery tendency to focus on the relative values of an acoustic input.

Conversely, temporal evolution of specific spectral energy bands has been captured by temporal patterns (TRAPS) and hidden activation TRAPS (HATS) \citep{Hermansky1994} features to detect underlying phonetic class structures usually taking long-time segments ($1$ second) compared to spectral analysis ($10$ ms). The hypothesis grounding the development of TRAPS and HATS suggest the spectral information perceived by the human auditory system serves not as classifier but as a frequency sub-band selector of the region most dominated by the target signal and thus temporal analysis of such bands is how the utterance is decoded in the cortex.

\citet{Kim2009}, proposed an algorithm to calculate power normalized cepstral coefficients (PNCC) as an alternative to the conventional MFCC. The calculation of PNCC integrates a Gammatone filterbank to better approximate the place-frequency mapping of the basilar membrane \citep{Patterson1992} as opposed to the triangular filters form MFCC, it also replaces their logarithmic non-linearity with a power function derived from physiological observations of auditory nerve firings to fit the dynamic dependency of the input sound level and the perceived loudness used to compress the output of the Gammatone filterbank; additionally, based on a ratio between arithmetic and geometric power mean PNCC are able to filter some of the background noise. A much broader overview of auditory-based feature extraction methods is exposed by \citet{Stern2012}.

Further physiologic and psychoacoustic research \citep{Qiu2003} \citep{Mesgarani2007} have shown the existence of neurons in the primary auditory cortex A1 of mammals specifically tuned to specific temporal or spectral modulations, and in some cases exhibit diagonal sensitivity patterns (such as vowel transients in speech). Spectro-temporal receptive fields (STRFs) are estimated patterns for time-frequency representations of stimuli optimally driving a neuron (or a group of neurons). To model such patterns, two-dimensional Gabor  filters were consequently developed to model patterns observed in STRFs \citep{Qiu2003}, owing to the localized spectro-temporal patterns explicitly coded in A1. \citet{Kleinschmidt2002} investigated if a set of those psychoacoustically parametrized filters, could extract meaningful information for robust ASR.

A challenge when designing filters for ASR is to determine a set of suitable parameters to produce a robust feature set able to deal with environmental noise, low signal-to-noise ratios, reverberation or even channel distortions. \citet{Schaedler2012} proposed a Gabor filterbank based on specific physiologically-motivated temporal and spectral modulation frequencies, which resulted in relative improvements of the word error rate (WER) by $30 - 45$\% compared to a MFCC baseline for ASR \citep{Meyer2012}, and $21$\% for speaker identification \citep{Lei2012}. 

In similar studies, a multitude of Gabor filters were employed to cover a wide range of modulation frequencies, and parsed as input to a large number of neural nets for merging the feature streams \citep{Zhao2008}. \citet{Ezzat2007}, based on 2D discrete cosine transforms, extracted spectro-temporal information to transform time-frequency patches of a spectrogram. 

Previously, we explored the applicability of Gabor filters arranged in a filterbank as input to DNN-HMM back-end on the Aurora\,4 task , which resulted in relative improvements of almost $20$\% over standardized filterbank features and $60$\% over MFCC results \citep{CastroMartinez2014}. Meanwhile, \citet{Chang2014}, using a different convolutional neural network initialized with a different set of Gabor filters, obtained fruitful results on the same task as well as in a re-noised version of wall street journal. Subsequently, \citet{Baby2015} proposed yet another auditory-based feature extraction method, which consist in low-frequency amplitude modulated spectrograms computed from low-passed-filtered half-way rectified signals; together with a DNN-HMM back-end, obtaining very similar WERs as we did for Aurora\,4 and $19.6$\% phone error rate on the TIMIT corpus. 

These studies support the idea of merging auditory-based features with deep learning architectures to get the best of both worlds, however, the cause behind the gain of such combination is still unknown. We believe the benefit comes from these representations which provide the deep learning decoder with more discriminable cues for the speech recognition task. Our aim with this paper is to validate this hypothesis by lowering the baseline word error rates (WER) in three different recognition tasks (Aurora\,4, CHiME\,2 and CHiME\,3) and assess the discriminability of the features encoded by Gabor filters. We pursue the latter objective analyzing the activations obtained from the DNN through a robust metric of separability in feature space. The indicated measure is the similarity between classes; being those the clustered context dependent triphone HMM states mapping to the same phoneme. 

The remainder of this paper is structured as follows: we describe in detail the Gabor filterbank, along with the baseline features, the setup of the deep neural network and the criteria used for the analysis in the \hyperref[methods]{methods} section. \hyperref[results]{Results} are presented in the following section, then a brief \hyperref[discussion]{discussion} depicting the results and the paper \hyperref[conclusions]{conclusions} afterwards.

\section{Methods}
\label{methods}

In this section we describe the auditory-based ASR features, i.e., Gabor features and the baseline filterbank features, the speech corpora, as well as the hybrid classification system which comprises a deep neural network and hidden Markov models. In the final part of this section, we present a method to assess the relevance of the auditory-based input streams for deep learning in comparison to the baseline features.

\subsection{Gabor Filterbank features}
\label{gabor}

\begin{figure*}[htb]
	\begin{minipage}{1.0\textwidth}
		 \includegraphics[width=\textwidth]{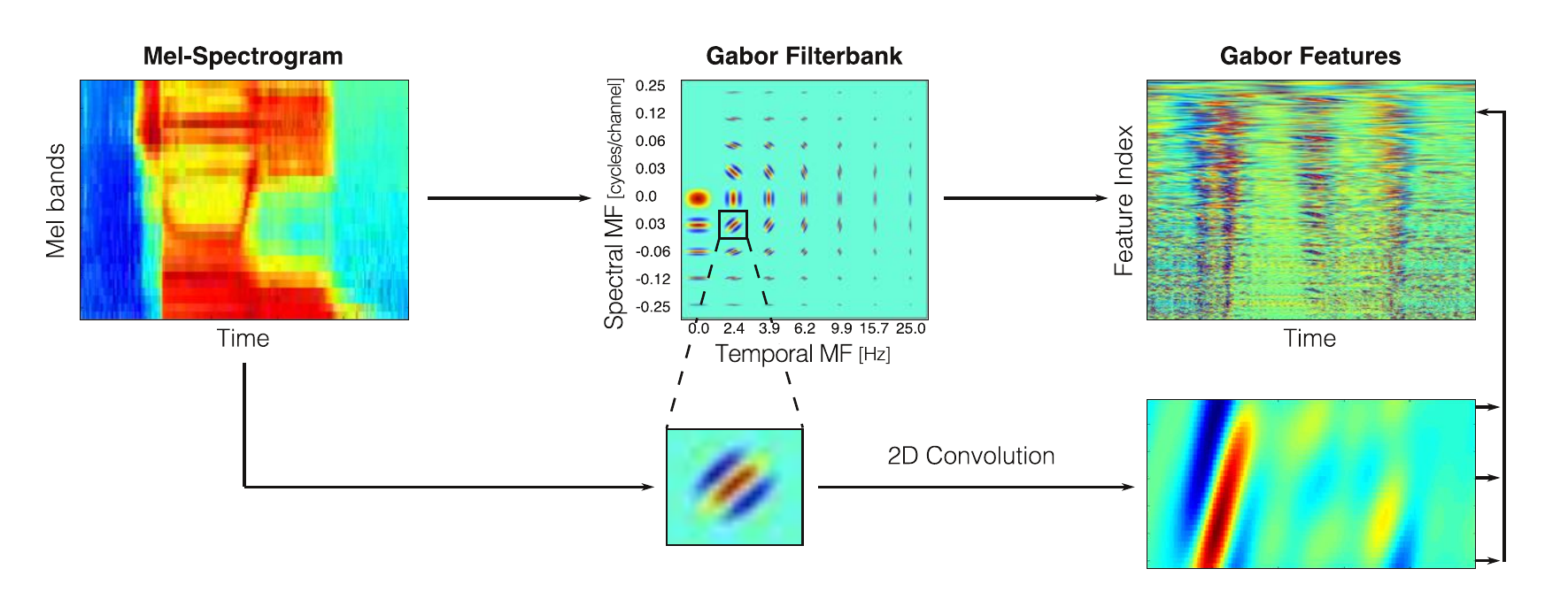}
    	 \vspace*{-1.8em}
	     \caption{Gabor filterbank feature extraction procedure. The input log Mel-spectrogram is convolved by each of the $59$ filters in the filterbank starting from the top and from left to right; the one taken as example below shows the contribution of this particular filter, the output is critically sampled and concatenated vertically giving the final 657-dimensional the complete feature vector representation.}
	     \label{fig:feature_extraction}
     \end{minipage}	  
\end{figure*}

Inspired by observations in spectro-temporal receptive fields in the auditory cortex (cf. previous section), we used a set of two-dimensional Gabor filters arranged in a filterbank to extract ASR features from speech signals. The procedure, depicted in \fref{fig:feature_extraction}, consists of three stages:

Initially, logarithmic Mel-spectrograms were extracted from the speech signals following the ETSI Distributed Speech Recognition Standard (\href{http://www.etsi.org/deliver/etsi_es/201100_201199/201108/01.01.03_60/es_201108v010103p.pdf}{\color{blue} 201 108 v1.1.3 2003}) with the only difference of using $31$ frequency channels instead of $23$. For the signals with a sampling frequency of $16$\,kHz, it provides a similar frequency resolution as the common $23$ channels for $8$\,kHz data employed, for instance, in the Aurora\,2 task \citep{Hirsch2000}.  Log-Mel-spectrograms were chosen as a starting point because they approximate the logarithmic compression of amplitudes and the non-linear frequency mapping of the auditory system. In the second stage, the spectrograms were convolved with every 2D filter in a modified version of the Gabor filterbank from \citep{Schaedler2012}. 

A Gabor filter is the product of a complex sinusoid function \eref{eqn:Gabor} and traditionally a Gaussian window; we replaced the latter with a Hann window \eref{eqn:Hann} to obtain better recognition scores due to better modulation frequency characteristics, as reported in \citep{Meyer2012}. The periodicity of the carrier sinusoid was defined by the radian frequencies $\omega_n$ and $\omega_k$ ($n$ and $k$ denoting time and frequency index, respectively), which allowed the Gabor filters to be tuned to particular spectro-temporal directions, as well as purely temporal ($\omega_k=0$) or purely spectral ($\omega_n=0$) modulations. 

The number of oscillations for the localized filters was kept constant for all filters, with a value of $3.5$ as suggested by \citet{Schaedler2012}. This procedure is similar to wavelet processing and would result in infinitely large filters for modulation frequencies of zero; hence, all filters were limited to a maximum size (in this case $69$ frequency channels and $99$ time frames). The envelope width was parametrized by the window lengths $W_n$ and $W_k$ and the center frequency channel $k_0$ and center time frame $n_0$.

\begin{align}\label{eqn:Gabor}
 \centering
	s(n,k) = \exp \bigl( i \omega_n (n-n_0) + \omega_k (k-k_0) \bigl)
\end{align}
\begin{align}\label{eqn:Hann}
 \centering
	h(n,k) = \frac{1}{4}\Biggr[1-\cos\biggr(\frac{2\pi(n-n_0)}{W_n +1}\biggl)\Biggl]\Biggr[1-\cos\biggr(\frac{2\pi(k-k_0)}{W_k +1}\biggl)\Biggl]
\end{align} 
\hspace{3em}

The Gabor filterbank contains a set of temporal, spectral and spectro-temporal filters to cover a wide range of modulation frequencies. Because frequency mapping is approximately linear at frequencies below $800$\,Hz rather than strictly logarithmic, spectral modulation frequencies are expressed in cycles per channel\footnote{Another unit could be cycles per mel, we opted for cycles per channel because it takes into account the mel scaling and also placing mel-frequency channels into bins. The mel-definition used in this work comes from the ETSI implementation calculated as follows: $Mel(x) = 2595 \log_{10}\biggr(1 + \frac{x}{700}\biggl)$}. The specific modulation frequencies were chosen so that the transfer functions of the filters exhibit a constant overlap in the modulation frequency domain. 

To account for modulations arising from syllable structure in spoken language, temporal modulations of $2.4$ and $3.9$\,Hz were included as in the slightly modified filterbank presented in \citep{Meyer2011} in addition to higher modulations also considered by \citet{Schaedler2012}. This resulted in 59 pairs of spectral and temporal modulation frequencies. 

With $59$ spectro-temporal filters and $31$ frequency channels, the resulting feature vectors would have been be rather high-dimensional ($1829$ components). However, filters with a large spectral extent produce highly correlated output between adjacent channels hence relatively small changes in the feature values when shifted by one frequency channel. Therefore, many frequency channels of larger filters were discarded from the feature matrix, while all channels were conserved for filters with the smallest spectral extent. This was achieved by choosing the channel centered on $1$\,kHz (which should contain information relevant for speech recognition) as well as channels obtained by shifting the current filter by one fourth of its spectral size and preserving its center frequency channel. Furthermore, as the Mel-spectrogram spectral size is smaller than the biggest Gabor filters, zero-padding was implemented to match the spectral content for the 2D convolution and to preserve the same number of features per frame without introducing significant boundary effects, the initial and last frame columns were padded on both temporal ends respectively.

The shifting value was selected based on the minimum window overlap needed for a perfect reconstruction of the spectrogram according to Nyquist-Shannon theorem. Alternative methods such as LDA and PCA were analyzed in \citet{Schaedler2012}. Critical sampling is designed to discard only redundant information, thus the number of selected channels lied between $1$ (for $\omega_k=0$ cycles/channel) and $31$ ($\omega_k = \bpm 0.25$ cycles/channel), and the feature dimension was reduced to $657$.\footnote{The original code used for feature extraction can be found in this repository: \href{https://github.com/m-r-s/reference-feature-extraction}{\color{blue} https://github.com/m-r-s/reference-feature-extraction}}

\subsection{Gabor Filterbank Subgroups}
\label{subgroups}

Given the wide range of spectral and temporal modulation frequencies taken into account in the Gabor filterbank, we became interested in knowing which particular set of filters is most relevant for DNN-based speech recognition. Hence, we divided the original filters into sets with low, medium and high temporal modulation frequencies, which resulting features are referred to as LTM (derived from filters with temporal modulations of $2.4$ and $3.9$\,Hz), MTM ($6.2$ and $9.9$\,Hz) and HTM ($15.7$ and $25$\,Hz), respectively.  Because critical sampling only removes spectral channels, all 3 subgroups are left with exactly the same channels.

\begin{figure}[htb]
	\begin{minipage}{1.0\linewidth}
		\includegraphics[width=\textwidth]{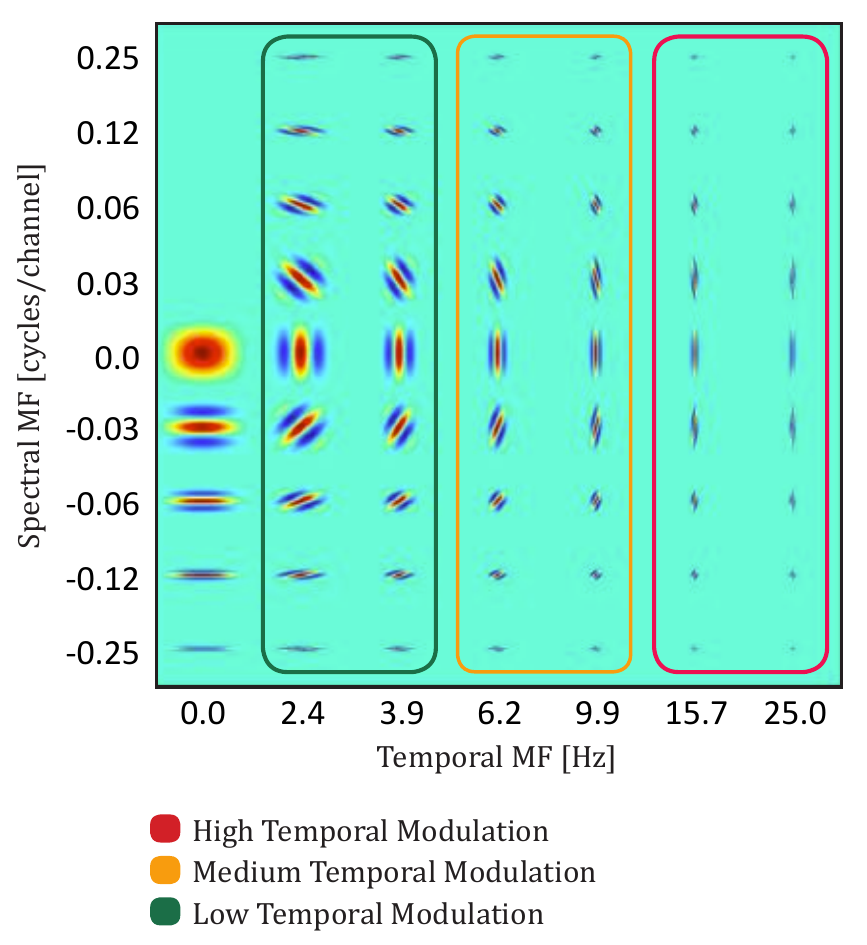}
	    \caption{Gabor filterbank subgroups.}
	    \label{fig:GBFB_subgroups}
    \end{minipage}	  
  	\centering
\end{figure}

Earlier experiments using a GMM-HMM recognizer and the Gabor filterbank indicated each individual 2D filter contributes to the noise robustness observed on the Aurora\,2 task \citep{Schaedler2012}, so we expected subgroups to perform worse than the full dataset. The filters selected from the complete set are highlighted in \fref{fig:GBFB_subgroups}. 

There were mainly two reasons for choosing this type of subdivision: first, our spectral modulations are given in cycles/channel and thus are more difficult to interpret and compare results than our temporal modulations in Hz. Secondly, lower temporal modulation frequencies have consistently been remarked as being most important in speech perception and recognition in the literature, which can be re-evaluated with this subdivision. It is important to mention how selecting a particular center frequency as temporal modulation does not necessarily exclude (only attenuates) the rest of the frequencies in the spectrum as the Gabor filters are broadband.

\citet{Kanedera1998,Kanedera1999} concluded the most useful linguistic information come from modulation frequencies components in the range of $2$ to $16$\,Hz (with $4$\,Hz as predominant component), and components above or below this range could degrade recognition accuracy. \citet{Drullman1994b,Drullman1994a} measured the perception of speech synthesized with several temporal envelopes for each frequency band and established the interval of modulation spectrum components between $4$\,Hz and $16$\,Hz to be the most critical for speech intelligibility; additionally, they reported a marginal contribution of modulation frequencies above $16$\,Hz when the lower components were present. Furthermore, the feature extraction procedure developed by \citet{Tchorz1999}  performed the best on temporal modulations around $6$\,Hz. All these studies provide strong reasons to expect  LTM to outperform HTM.

Owing to the critical sampling, the feature vectors extracted with the Gabor filterbank do not contain an equal amount of channels from each filter, i.e. the number of bands produced by the convolution of the log-Mel-spectrogram with each filter depends on the size of the filter, however, as each aforementioned subgroup includes all the spectral modulation frequencies, the dimension of output is $202$ for every subgroup. 

\subsection{Baseline}
\label{mfsc}

Raw Mel-Filterbank features have been found to outperform MFCC features in recognizers with DNN-based architectures and hence serve as baseline features \citep{Mohamed2012}: Log-mel-spectral coefficients (MFSC) are obtained from the $31$ channels of the same spectrogram used for calculating the Gabor features. As these filterbank representations include more information of the original Mel-spectrogram than MFCCs (where only the first $13$ bands are selected), a DNN is less constrained to create any structure of the input data; thus provides us a better contrast for our "hand-crafted" features.

A conventional triphone GMM-HMM recognizer was built prior to the deep neural network in order to obtain target labels via force-alignment. Per speaker a single feature-space maximum likelihood linear regression transform was calculated to train this model adaptively.

%-----------------------------------
%	SECTION DNN
%-----------------------------------

\subsection{DNN implementation}
\label{dnn}

The deep neural network (DNN) was based on the one described in \citep{Vesely2013}. Recognition experiments were conducted using the Kaldi ASR toolkit \citep{Povey2011}. Due to the capability of unsupervised pre-training using Restricted Boltzmann Machines (RBM), which provides a deep hierarchical representation of the training data, we opted for Karel's recipe.

In principle, this implementation can be summarized in two phases: pre-training and cross-entropy tuning. On the former phase, a stack of RBMs, also known as a deep belief network (DBN) \citep{Mohamed2012}, was trained in a greedy fashion one layer at a time using contrastive divergence as described by \citet{Hinton2010}. 

On the latter phase, serving as a backbone for the final network, the DBN was fine-tuned to classify frames into triphone-states using an independent (development) set and the cross entropy between the network output and the labels as a cost function.

The training was done in up to $20$ epochs (stopping when the relative improvement was lower than $0.001$). The starting learning rate was $0.008$ (halving it every time the relative improvement was lower than $0.01$) and no momentum nor regularization techniques (such as $L1 \& L2$) were applied. A soft-max layer of approximately $2000$ units was attached to the end of the DNN to output the most likely posterior probabilities of each context-dependent HMM state.  

The resulting DNN had $2048$ sigmoid neurons on each of the six hidden layers. Optimization via stochastic gradient descent was performed on a graphics processing units for speeding purposes. The size of the input layer varied depending on the type of the feature, for any given one $11$ frames are spliced to provide a context of $\bpm 5$ frames. 

In a nutshell, for every feature a GMM system was trained without changing any baseline configurations (except for the feature themselves) to provide the alignment of context dependent states to frames; then pre-training is performed to initialize the DNN, which uses the class labels provided by the GMM system; after fine-tuning the DNN is retrained, only this time it uses the labels produced by the DNN instead.

\subsection{Discriminability Criteria} 
\label{similarity}

Usually ASR Systems are evaluated in terms of the word error rate over a testing set. As we wanted to have a better understanding of the particular relationship between deep learning and auditory-based features we decided to observe the activations from the DNN output layer instead of analyzing the features separately. 

Phoneme discriminability characterizes the performance of the learned representations even if the DNN is trained to deliver scaled probabilities of the senone HMM states, because each transition can be seen as a branch of a correspondent decision tree. The roots of those trees are the central phoneme of the trained triphones and are used to create phoneme classes from the clustered branches.

We selected a list of phonemes in the ARPA format (ARPABET) and gathered a group of activations corresponding to only the frames labeled as the phonemes in this list. The corpora used for this analysis had a disparate number of examples for each phone, so we created separate lists, one for the large vocabulary tasks (i.e, Aurora\,4 and CHiME\,3) and a different one for CHiME\,2.

Labels were taken from the clean sets when possible to ensure they convey the spoken message; for Aurora\,4 using the given clean close-microphone condition; for CHiME\,3, as there are no available clean labels, we used the ones produced by the force-alignment from the best performing setup; the CHiME\,2 clean labels were used from forced alignment system detailed in \citep{Kabir2010}. 

Owing to the high dimensionality of the activations, a measure of separability robust to reparametrization was needed. As a criterion for assessing how well a particular feature separates the input into distinguishable classes we chose the cosine similarity. For being $L_2$ based, this metric is invariant to rotation of the coordinate system and thus allowed us to compare the discriminability among the different features. The cosine similarity is defined as:

\begin{align}\label{eqn:Similarity}
	S(\vec{v_1},\vec{v_2}) = \frac{\vec{v_1} \cdot \vec{v_2}}{\|\vec{v_1}\| \|\vec{v_2}\|}
\end{align}

Where each vector is the centroid of all the gathered examples for a given phoneme class (mean and variance normalized), the numerator is the inner product between the correspondent phoneme classes and the denominator is the product of their norm. The cosine similarity measures the relationship between two vectors represented by a value between $0$ and $1$; the closer this value gets to $0$ the wider the angle formed by these two vectors. 

Generally, the similarity represents how close the phoneme manifolds are projected in the hyperspace, therefore a higher value increases the difficulty for phoneme separation performed by the DNN. Classes with a larger distance are less likely to be confused, conversely smaller angles (similarity values close to $1$) lead to higher misclassifications. By calculating the similarities between every phoneme in the list with each other, we obtained the similarity matrices shown in the following section. 

These matrices would ideally be identity matrices, so the more a similarity matrix resembles an identity matrix the better the classification capabilities of the system. For each corpus a similarity matrix was calculated to get a better understanding of the relevance of the information encoded by the auditory-based features in combination with the DNN for the recognition task.

\subsection{Corpora}
\label{corpora}

We performed a series of speech recognition experiments using different Corpora to prevent the system configuration and post-analysis from being adapted to a particular task. The Aurora\,4 Corpus and the one used for the 3rd CHiME (Computational Hearing in Multisource Environments) Challenge are both derived from the same large vocabulary continuous speech recognition task, namely Wall Street Journal \citep{Garofalo2007}; the latter, however, also contains real recordings in noisy environments which provides more realistic data for the analysis. 

The small vocabulary GRID-based corpus originated from the 2nd CHiME Challenge was of particular interest for us because the test set is available in multiple signal-to-noise ratios (SNRs) and therefore enables an analysis across different noise levels, furthermore, being a short vocabulary task allow to detect if there is a particular effect from the vocabulary size or the language model.

\subsubsection{Aurora4} 
\label{aurora}

The Aurora\,4 framework \citep{Parihar2004} was used to assess the impact of additive noise from different sources and the effect of channel distortions; it is a large vocabulary continuous speech recognition task derived from the standard LDC Wall Street Journal (WSJ0) corpus. We opted for the multicondition set for training, which consists of $7137$ utterances from $83$ independent speakers, one half of the $16$\,kHz files were recorded with the close-talk Sennheiser HMD-414 microphone, the other half using one of $18$ different types of microphones. 

Each half was further subdivided; no noise was added to one fourth ($893$ utterances) while the remaining three-fourths ($2676$ utterances) were corrupted with one of six different types of noise (car, babble, restaurant, street, airport and train) at randomly selected SNR conditions between $10$ and $20$ dB. The test set included in the framework was extracted from the WSJ0 $5,000$ word closed-vocabulary task which consists of $330$ utterances from $8$ speakers repeated in the same $14$ conditions used in the train set at $5$ to $15$ dB SNR. 

\subsubsection{CHiME3}
\label{chime3}

For the third CHiME Challenge \citep{Barker2015}, sentences contained in the WSJ0 corpus were recorded using a 6-microphone tablet in four \textit{real-life} scenarios: café, street junction, public transport and pedestrian area. Additionally, the task includes also \textit{simulated} noisy utterances to assess the value of generated data, as it is easier and cheaper to obtain and could be potentially useful for training purposes.

A total of $12$ US English speakers ($6$ male and $6$ female) ranging in age from $20$ to $50$ years were recorded after short test sessions to ensure each speaker performed the reading task correctly. An interface showed the talkers approximately $100$ sentences to be read; those were recorded in an isolated booth (which served as basis for the simulated data) and in each location as described above. The training set comprised $1600$ real noisy utterances ($4$ speakers x $100$ sentences x $4$ scenarios), whereas the development and test set consist of the same $410$ and $330$ utterances from the WSJ0 corpus, randomly divided in $4$ subgroups and read on each scenario, resulting in $1640$ and $1320$ utterances respectively.

We used only the "noisy" set (utterances from the frontal closest microphone Channel $5$) for training and testing to exclude from the analysis uncontrolled gains as a result of the speech-enhancement technique. 

\subsubsection{CHiME2-GRID}
\label{chime2}

The CHIME2-GRID dataset \citep{Vincent2013} from the second CHiME challenge was included for two main reasons: to have an estimate of the class separation performance from auditory based features and deep learning over different SNR levels and to verify if the proposed set up performed well on small vocabulary tasks. The GRID corpus \citep{Cooke2006a}, from which this data was extracted, consist of 6-word sequences read by $34$ speakers of the form: $<$command:4$>$ $<$color:4$>$ $<$prepos.:4$>$ $<$letter:25$>$ $<$digit:10$>$ $<$adverb:4$>$, (e.g. "bin green at C 5 now") where the numbers in brackets indicate the number of choices per word. 

These utterances were generated using binaural noise recordings from a head and torso simulator in a living room and mixed with the GRID data at $6$ different SNR levels from $-6$\,dB to $9$\,dB in steps of $3$\,dB. Moreover, each utterance was convolved with a set of head impulse responses simulating speaker movements and reverberation to make the task more realistic. We used the isolated noisy $16$\,kHz $500$ utterances from each of $34$ speakers as the training set, and the $600$ utterances at each of the $6$ SNR conditions as test sets.

\section{Results}
\label{results}

We experimentally confirm the effectiveness of auditory-based Gabor features across three different speech recognition tasks. The performance in terms of word error rate (WER) for four Gabor feature sets (full filterbank and the three subgroups according to the temporal modulation frequencies) and the filterbank baseline is presented in \tref{tab:WER}

For simplicity the $14$ conditions from Aurora\,4 were grouped into $4$ subsets: "A" and "B" correspond to the clean and noisy recorded using the close-talk microphone, respectively, and likewise "C" and "D" for the clean and noisy utterances with different channel characteristics introduced by the different secondary microphones. The CHiME\,3 rows come from the real-recordings (real) and the simulated data (simu) parts of the test set. The bottom rows are the WER from the CHiME\,2 test set, the number in parentheses indicates the corresponding SNR.

The features obtained from Gabor filters with low temporal modulations (LTM) produce notably the worse results in all three tasks. Conversely, the representations derived from filters from high temporal modulations (HTM) perform consistently better than all the others in each task. The second to best features for all conditions, except Aurora\,4 "A", are the ones extracted from medium temporal modulation Gabor filters (MTM), whereas the complete Gabor filterbank (Gabor) produced features robust in noisy environments but not as good as the ones from raw filterbank (MFSC) in cleaner scenarios.   

\begin{table}[htb]
\renewcommand{\arraystretch}{1.4}
\centering
\resizebox{\columnwidth}{!}
{%
\begin{tabular}{l c c c c c} 
 \toprule
 \toprule
& MFSC & Gabor & LTM & MTM & HTM \\
 \hline
Aurora\,4 (A) & 3.9 & 3.9 & 12.4 & 4.6 & 3.0 \\
Aurora\,4 (B) & 7.5 & 8.5 & 21.2 & 7.8 & 5.6 \\ 
Aurora\,4 (C) & 12.3 & 8.8 & 20.4 & 9.0 & 6.3 \\ 
Aurora\,4 (D) & 22.2 & 19.2 & 34.6 & 18.4 &	15.4 \\ 
 \hline
CHiME\,3 (simu) & 23.3 & 30.8 & 41.6 & 20.2 & 15.2 \\ 
CHiME\,3 (real) & 36.0 & 40.1 & 50.8 & 26.6 & 21.0 \\ 
 \hline
CHiME\,2 (9dB) & 4.8 & 6.1 & 7.3 & 4.9 & 4.4 \\       
CHiME\,2 (0dB) & 12.6 & 12.3 & 15.4 & 11.9 & 10.7 \\  
CHiME\,2 (-6dB) & 26.0 & 20.3 & 25.3 & 20.6 & 19.9 \\ 
 \bottomrule
 \bottomrule
\end{tabular}
}
\caption{Word Error Rates for the three speech recognition tasks comparing the baseline and the auditory-based Gabor features processed with Deep Neural Networks}
\label{tab:WER}
\end{table}

The whole Gabor filterbank yields similar results as MFSC on conditions "A" and "B" of the Aurora\,4 framework; on conditions "C" and "D" the effect of different channel characteristics can be appreciated as the WER decreases drastically even in the absence of additive noise. The WER increase from condition "C" to "D" is almost uniform ($9 - 10$\%) for all features except LTM and is considerably larger than the one from "A" to "B". 

Concerning the CHiME\,3 task, the WER difference between the \textit{real} and the \textit{simulated} test sets can be used to quantify the generalization capabilities of the systems trained on different features because the generated data do not entirely capture the acoustic complexity of a \textit{real-life} scenario. For MFSC features this difference is the biggest. 

\begin{figure*}[htb!]
	\begin{minipage}{1.0\textwidth}
		 \includegraphics[trim=0.7cm 9.4cm 0.6cm 1.5cm, width=\textwidth]{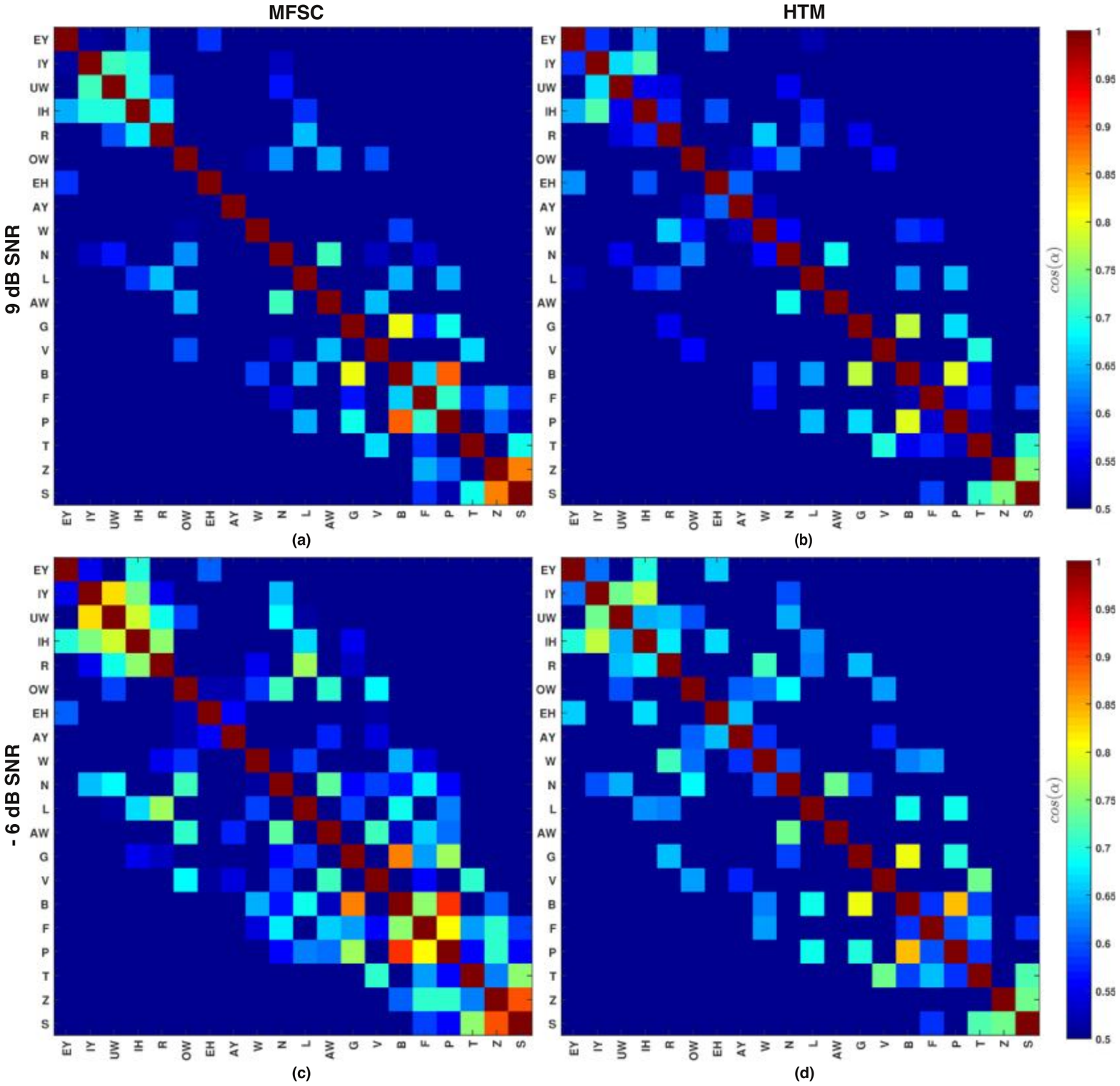}
    	 \vspace*{-0.5em}
	     \caption{Similarity matrices of MFSC (left) and HTM (right) from the CHiME\,2 Corpus. On the upper row the SNR is $9$\,dB and $-6$\,dB on the lower row. For readability a lower threshold was set to $\cos(60^{\circ})$ so that angles wider than $60^{\circ}$ would be ignored.}
	     \label{fig:CHiME2_SM}
     \end{minipage}	  
     \centering
\end{figure*}

For the small vocabulary task, auditory-based Gabor features show robustness against low SNRs (except LTM); for Gabor and MTM this advantage is revealed when lowering the SNR below $0$\,dB, whereas HTM yield lower WERs even at positive levels. MTM perform almost the same as MFSC at $9$\,dB and $0$\,dB The actual assignment evaluated in the first track of the CHiME\,2 Challenge was to correctly recognize the letter and digit tokens. To relate the recognition scores to the post-analysis shown in \fref{fig:CHiME2_SM}, where more phoneme samples were needed, we based our WER scores on the whole utterances. 

To test whether the low temporal modulation filters are the culprits for the higher WER, we decided to restrict network-related optimization effects with the following configurations on Aurora\,4: a) LTM + HTM (referred to as LHTM);  b) MTM + HTM (MHTM); c) features produced by DC filters + HTM (DCHTM), the former being the filters with temporal modulation frequency of $0$; d) random noise + HTM (RHTM) and e) a matrix of 0's + HTM (ZHTM). The last two set-ups contain either uniformly distributed numbers in the range of [$-1$ - $1$] or the an equal number of zeros, both matching the size of HTM, thus making all configurations but the third one equidimensional. Average WERs are shown in  \tref{tab:Aurora4extraWER}.

\begin{table}[htb]
\renewcommand{\arraystretch}{1.4}
\centering
\resizebox{\columnwidth}{!}
{%
\begin{tabular}{l c c c c c c} 
 \toprule
 \toprule
& LHTM & MHTM & DCHTM & RHTM & ZHTM & HTM\\
 \hline
Aurora\,4 & 13.10 & 11.55 & 11.35 & 11.02 & 10.75 & 9.66 \\ 
 \bottomrule
 \bottomrule
\end{tabular}
}
\caption{Average Word Error Rates for the Aurora\,4 task comparing HTM and 5 additional configurations: LTM + HTM (LHTM), MTM + HTM (MHTM), features produced by DC filters + HTM (DCHTM), random noise + HTM (RHTM) and a matrix of 0's + HTM (ZHTM)}
\label{tab:Aurora4extraWER}
\end{table}

HTM outperformed the remaining configurations. Adding LTM  drastically increases the error, more so than MTM, features from DC filters, zeros or even 0's. Random noise is slightly more detrimental than 0's as a complement to HTM; likewise MHTM performs worse than DCHTM. Only average results are shown as the same trend is observed consistently over each condition shown in \tref{tab:WER}.

\begin{figure*}[htb]
	\begin{minipage}{1.0\textwidth}
		 \includegraphics[trim=0.7cm 19.4cm 0.6cm 1.5cm, width=\textwidth]{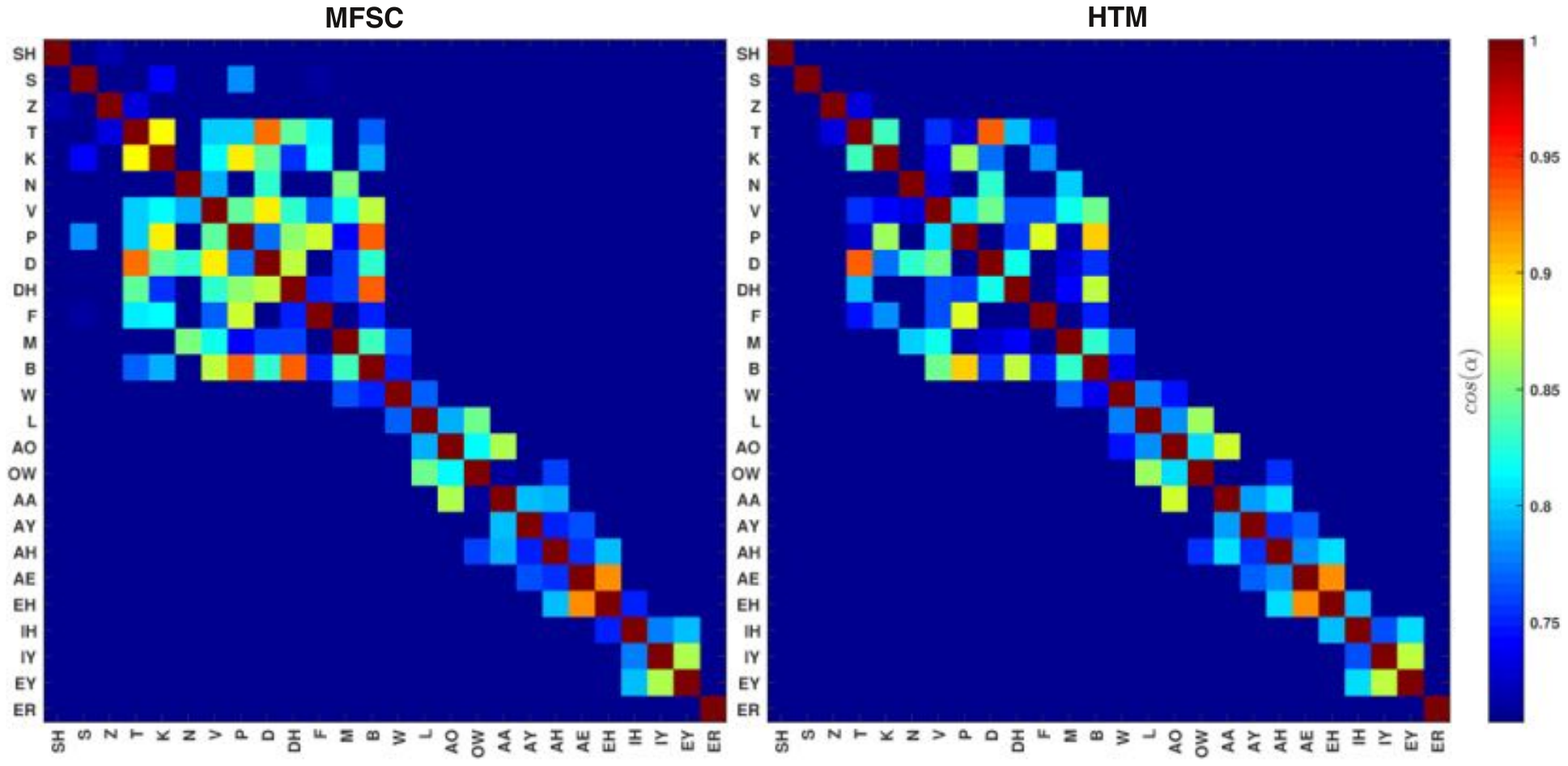}
    	 \vspace*{-0.5em}
	     \caption{Similarity matrices of MFSC (left) and HTM (right) from the Aurora\,4 corpus. For readability a lower threshold was set to $\cos(45^{\circ})$ so that angles wider than $45^{\circ}$ would be ignored.}
	     \label{fig:Aurora4_SM}
     \end{minipage}	  
     \centering
\end{figure*}  

Owing to the performance of HTM, a post-analysis was implemented using the discriminability criterion described in section \ref{similarity}. \fref{fig:Aurora4_SM} shows the similarity matrices of MFSC and HTM on the Aurora\,4 task. In order to reduce the bandwidth of the matrices, the phonemes were arranged through an implementation of the sparse reverse Cuthill-McKee ordering \citep{Gilbert1992}. For readability a threshold of $\cos(45^{\circ})$ was set in order to filter out the angles wider than $45^{\circ}$; thus the non-zero elements represent similarity values above $0.7$, which are likely candidates for phoneme confusions. A clear distinction is remarked between consonants on the upper left corner and vowels on the lower right. Both features produced almost the same similarity patterns for vowels, in this corner, the most noticeable confusions are between phones: /EY/ \& /IY/, /AA/ \& /AO/, /AE/ \& /EH/ and /L/ \& /OW/.

For consonants MFSC produced more confusions than HTM. The highest similarities between 2 phonemes occur among cases: /T/ \& /D/ for both features; the values on cases /P/ \& /B/, /DH/ \& /B/, /T/ \& /K/, /P/ \& /K/ and /M/ \& /N/ are predominantly higher for MFSC. A pattern of multiple confusions appears on the MFSC figure for phones: /V/, /D/ and /DH/ with respect to /T/, /K/, /N/, /P/ and /B/; whereas on HTM figure the same pattern has almost vanished (except for /D/ \& /T/ and /DH/ \& /B/). Phonemes /P/ \& /F/ yield almost the same similarity for both features.

\fref{fig:CHiME2_SM} shows the similarity matrices of MFSC on the left side and HTM on the right side calculated from the CHiME\,2 framework. The upper row containing figures (a) and (b) were computed using the test set at $9$\,dB SNR and $-6$\,dB on lower row with figures (c) and (d). This time the threshold was set at $60^{\circ}$ as the number of phonemes were reduced, thus the matrix elements represent values above the similarity value of $0.5$. Nevertheless, we still refer to elements whose similarity value is above $0.7$ as confusions.

For MFSC the similarity between the phoneme classes /Z/ \& /S/ is high even at $9$\,dB and greater than the one of HTM. There is also a strong similarity of phonemes /B/ \& /G/ for both features, but in \fref{fig:CHiME2_SM}\,(c) the value gets closer to $1$. With respect to phoneme /F/ there are high similarities with phonemes /B/ and /P/ for MFSC features whereas these values are below threshold in \fref{fig:CHiME2_SM}\,(b) and almost so in \fref{fig:CHiME2_SM}\,(d) for HTM. 

At $-6$\,dB SNR several confusions appear on MFSC among the pairs of phonemes /R/ \& /L/, /IH/ \& /UW/ and /IH/ \& /R/ and on HTM between the phonemes /R/ \& /W/. In  \fref{fig:CHiME2_SM}\,(c) a \textit{noisy} pattern, comparable to the one formed on phonemes /V/, /D/ and /DH/ in \fref{fig:Aurora4_SM} for MFSC, can be observed forming on phoneme /F/ with respect to almost every phoneme from /N/ to /S/ (except phoneme /L/); this pattern is once again diminished for HTM (\fref{fig:CHiME2_SM}\,(d)).

Another noticeable likely confusion among the four graphs is the one between /B/ \& /P/; for both conditions (i.e. $9$\,dB and $-6$\,dB), however, the similarity increased twice as much for MFSC from $0.86$ to $0.96$ compared to HTM in which case the value went from $0.80$ to $0.85$. A similar pattern occurred in phonemes /IY/ \& /UW/ and /T/ \& /S/. Conversely, between phonemes /IY/ \& /IH/ and /N/ \& /AW/ the same deterioration of approximately $0.05$ units was observed on both features when the SNR lowered $15$\,dB. 

To further explain the robustness of HTM an extra set of experiments was conducted: firstly, for the most challenging scenarios in Aurora 4 (for instance, using a secondary microphone in a train station additive noise), both the features and the activations were analyzed to inspect saliency of features for specific speech sounds and to review the resulting performance in terms of phoneme classification on basis of the posteriorgram. 

\begin{figure*}[htb!]
	\begin{minipage}{1.0\textwidth}
		 \includegraphics[trim=1.5cm 20.0cm 1.8cm 1.5cm, width=\textwidth]{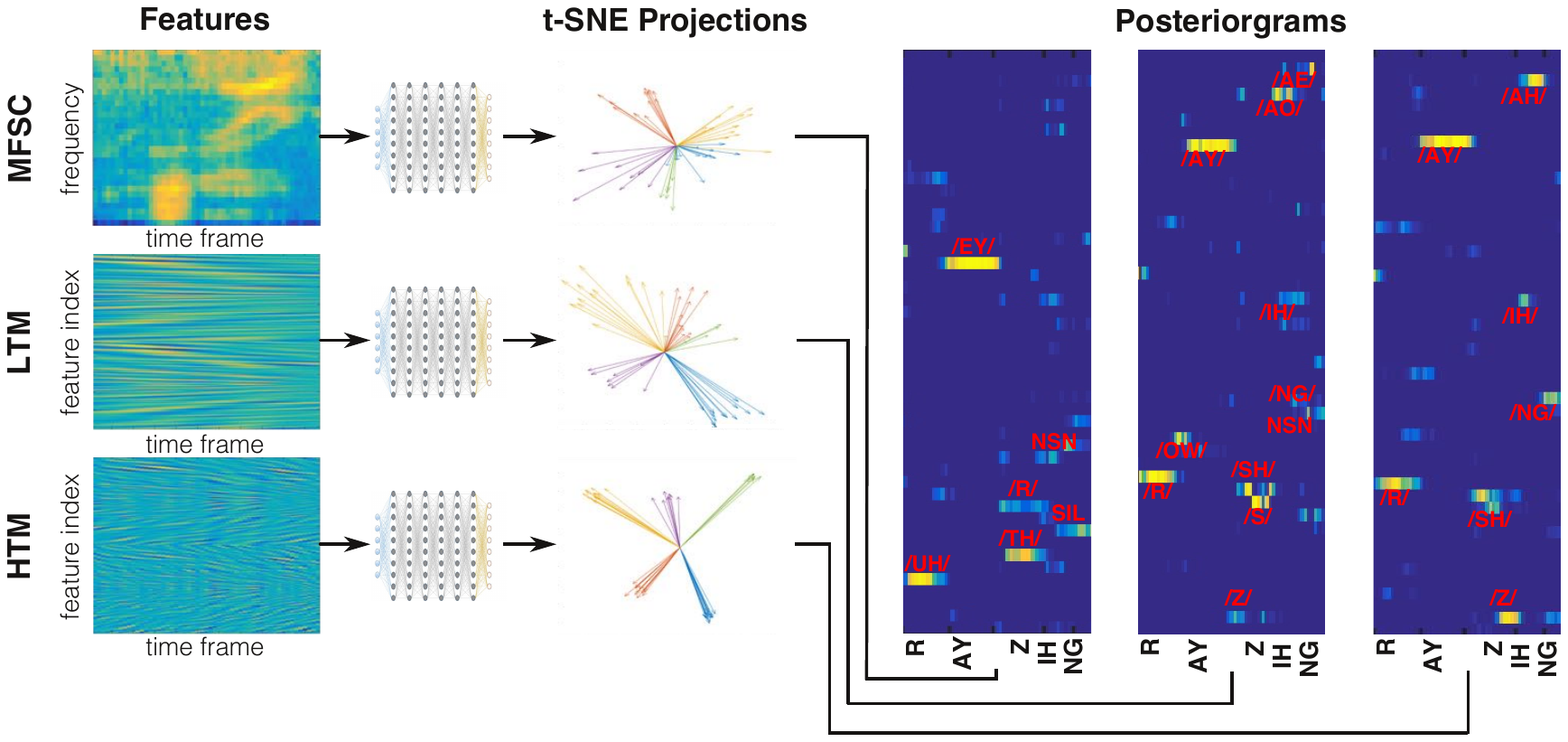}
    	 \vspace*{-0.5em}
	     \caption{Example of the word \textit{raising} extracted from one of the most challenging Aurora\,4 scenarios (restaurant noise with secondary microphone). The first column are the features of this 53-frame segment, the t-SNE projections in the middle representing the discriminability per frame in the 5 phoneme classes, finally the resulting posteriorgram highlighting the strength of the activations produced by the DNN.}
	     \label{fig:PGram_Aurora4}
     \end{minipage}	  
     \centering
\end{figure*}

We found high temporal modulations produced a more confident decision (based on the activation strength) of the accurate label. \fref{fig:PGram_Aurora4} exemplifies this analysis condensing the evaluation performed on the activations under the worst performing scenarios. To analyze the structure of the projected classes from a trained setup, we recurred to the visualization technique called t-distributed stochastic neighbor embedding (t-SNE) proposed by \citet{Maaten2008}, which allowed us to observe the distribution of the target classes in low-dimensional manifolds.

\begin{figure}[htb!]
	\begin{minipage}{1.0\columnwidth}
		 \includegraphics[trim=0.1cm 0.6cm 0cm 0.2cm, width=\columnwidth]{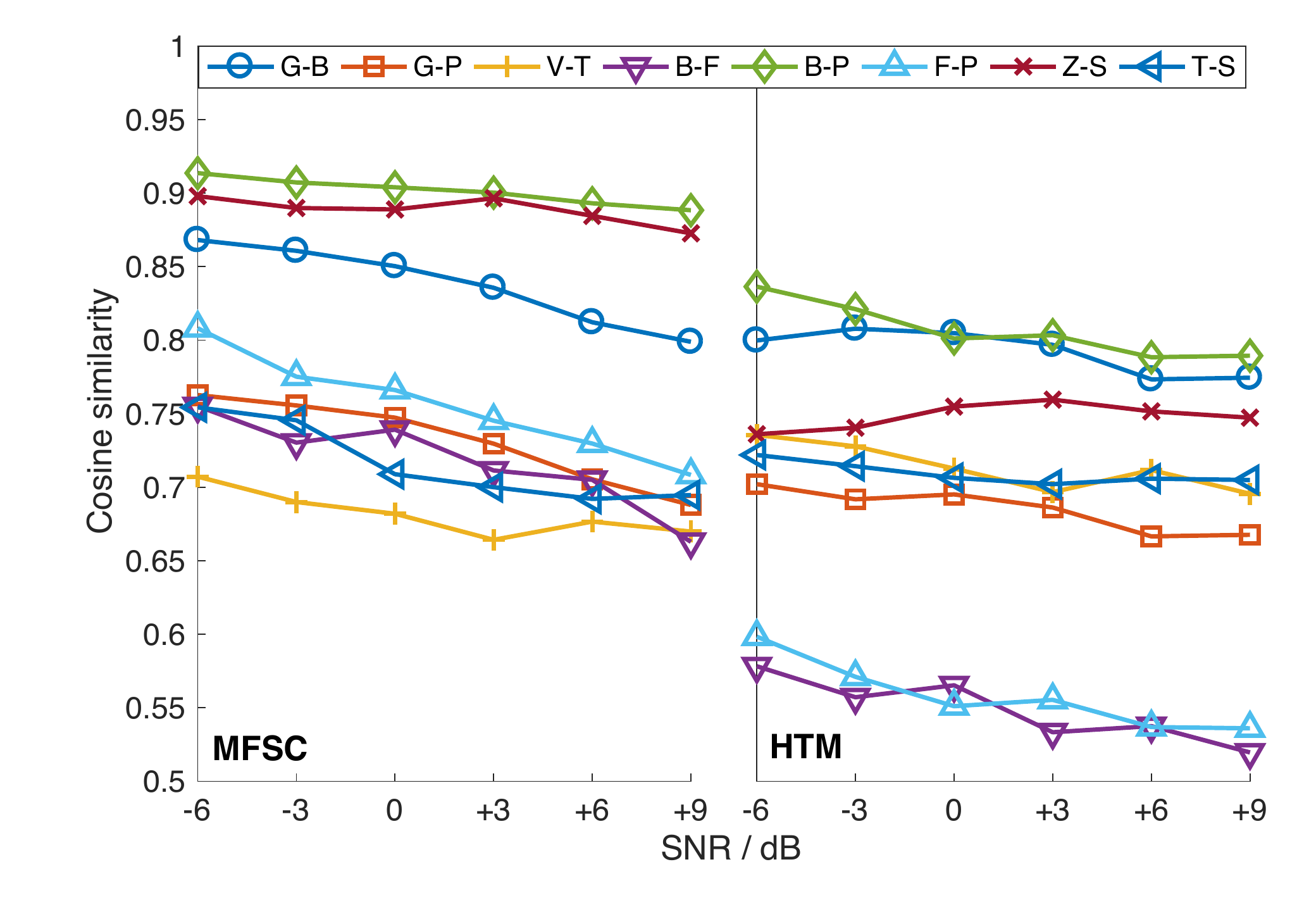}
    	 \vspace*{-0.5em}
	     \caption{Confusion Patterns for the most relevant pairs pf phonemes from the CHiME\,2 task for MFSC and HTM at different SNR conditions.}
	     \label{fig:Confusion Patterns}
     \end{minipage}
     \centering
\end{figure}

Secondly, we gathered main confusion patterns from the activations produced by all features in the CHiME\,2 task. \fref{fig:Confusion Patterns} summarizes the most relevant confusion patterns on HTM and MFSC over all SNR conditions of the test set. Overall, the similarity value decreases when the SNR increases which supports the idea of more separable projections leads to better recognition scores. 

Among the phonemes studied, the most confusable patterns were /B/ \& /P/, closely followed by /Z/ \& /S/ for MFSC and /G/ \& /B/ for HTM, peaking at a highest similarity of $0.83$ for HTM and $0.91$ for MFSC corresponding to an estimated minimal separation between classes of approximately $33^{\circ}$ and $24^{\circ}$ respectively. The following pairs seem to be equally challenging for both features: /T/ \& /S/ and to a lesser extent, /G/ \& /B/ and /G/ \& /P/. The pair of phonemes /V/ \& /T/ is clearly more discernible for MFSC; the opposite trend is observed in the case of /B/ \& /P/, /Z/ \& /S/, /B/ \& /F/ and /F/ \& /P/.

\section{Discussion} 
\label{discussion}

The results presented in \tref{tab:WER} clearly show how exploiting a particular set of Gabor filters with high temporal modulation frequencies in combination with a deep neural network provides a boost in performance on three different recognition tasks. HTM yielded the lowest error rates, even in clean conditions where the baseline MFSC achieve already a high accuracy.

\citet{Meyer2010} used an stochastic approach to determine Gabor filter parameters relevant for ASR and found positive contributions for a wide range of temporal modulations frequencies from $2$ to $25$\,Hz. Similarly, RASTA-PLP features, a range of $2.6 - 20$\,Hz was found to be useful. Hence, we expected the subgroups with a limited modulation range to perform worse than the complete Gabor filterbank, especially because fully connected DNNs are capable of handling correlated signals more effectively than traditional models. The results, however, show otherwise; both MTM and HTM outperform the whole filterbank. 

Moreover, the inclusion of lower temporal modulation frequencies ($2 - 4$\,Hz) seems to severely harm the representations extracted by the Gabor filterbank as LTM features yield the highest WER in all conditions and corpora. This is in contrast to previous studies \citet{Kanedera1998,Kanedera1999}, \citet{Drullman1994b,Drullman1994a} and \citet{Tchorz1999} that indicate high temporal modulation frequencies (above $16$\,Hz) to deteriorate performance on speech related tasks, from perception, recognition and intelligibility. \citet{Ganapathy2014} arrived to a similar conclusion, their band-pass filtering results suggest $15$\,Hz is an optimal upper cut-off limit for speech recognition performance in noisy conditions.

\citet{Kanedera1998} found $4$\,Hz to be the dominant component encoding the most useful linguistic information. In contrast to these contributions, the proposed high temporal modulations for Gabor processing do not filter out sharply contiguous regions among the spectra; by design the Gabor filterbank contain constant-Q filters, therefore their bandwidth is proportional to the center modulation frequency. 

In other words the higher the modulation frequency the broader the bandwidth. So even both subgroups of filters reach contiguous frequencies, the gain of the individual transfer function is higher in neighboring frequencies for filters with higher central modulation frequency\footnote{For more details about the individual gain of each filter we refer the reader to \citep{Schaedler2012}}. 

We argue one of the reasons focusing on $16$ and $25$\,Hz as a center frequency boosts recognition performance is because the produced features mimic the important strategy found in human listening to rely on localized patterns in the time-frequency representation (\textit{glimpsing}), as pointed out in \citep{Cooke2006b}, where the target signal dominates the noise, hence speech-relevant information can be extracted from the glimpses encoded in HTM.

The size of filters presumably also plays an important role: Given that low temporal modulation filters have the largest temporal extent in the Gabor filterbank, they produce stronger temporal smearing which might prevent the DNN from extracting phoneme-specific patterns. Neurons in the first layer compute a linear function of the input; feeding them with shorter segments of the spectrogram (i.e. smaller filters) allows higher layers containing non-linearities to learn more sparse and distributed features, thus resulting in fewer confusions as discussed below.

Recently, \citet{Chait2015} conducted experiments to support the idea of multi-time resolution processing taking place in human speech perception. Their findings disprove low temporal modulations as sufficient for speech recognition and remark the need of a model to include higher modulation frequencies as well. Such a combination has been tested (on a GMM-HMM recognizer) by \citet{Hermansky2005} extending the RASTA processing (mentioned in Section \ref{intro}) with a bank of two-dimensional filters to incorporate temporal trajectories of critical-band spectrograms; this step is followed by a TANDEM feature extraction. Albeit an akin approach to feature extraction with the complete Gabor filterbank; in this work, we found higher temporal modulations alone the most beneficial for ASR.

Each recognition task evaluated different aspects, for instance, Aurora\,4 highlights the effect of additive noise and channel distortions at positive SNR levels. The changes in WER between conditions "A" and "B" as well as "C" and "D" represent the effect of additive noise. In the same way the detriment from "A" to "C" and "B" to "D" measures the effect of different channel characteristics. The complete Gabor set appears to be more robust against channel distortions compared to MFSC with consistent improvements for test sets C and D. This robustness is preserved for HTM plus a higher robustness against additive noise with an average relative improvement of $29$\% is achieved over the MFSC baseline. 

Among the features shown in \tref{tab:Aurora4extraWER}, HTM has the lowest number of feature components and yielded the lowest WER, suggesting there is not a significant effect in the recognition scores due to dimensionality. Concatenating random noise to HTM decreases performance slightly more than just adding zeros which reflects the capacity of the network to ignore uninformative input. 

Surprisingly, the combination of HTM and the features extracted from DC filters resulted into lower WER than when combining MTM and HTM, even though the former was the second best feature from the ones compare in \tref{tab:WER}. We assume the MHTM combination contains more redundant components than DCHTM, which should have a detrimental effect.

Several successful approaches have been reported on the Aurora\,4 task focusing on improved speech enhancement or feature extraction in deep learning systems: For instance, an exemplar-based speech enhancement proposed by \citet{Baby2015} resulted in a WER of $11.9$\%. \citet{Chang2014} investigated Gabor filters in convolutional deep neural networks and obtained a $16.6$\% WER. 
 
Similarly, improved net architectures have been investigated: \citep{Rennie2014} trained an order statistic network with an \textit{annealed} version of the dropout regularization method obtaining $10.0$\% WER. \citet{Geiger2014} achieved a $13.3$\% WER by implementing a long short-term memory recurrent neural network (for its ability to exploit temporal context) in combination with a non-negative matrix factorization for speech enhancement. \citet{Mitra2014} worked on both approaches and switched the fully connected deep neural network for one with convolutional layers together with vocal tract length normalization and lowered the WER on all conditions using a uniformly weighted combination of $5$ acoustic features (WER: $14.1$\%). 

With a relatively simple approach of replacing the feature extraction, a WER of $9.7$\% was obtained in this study, which potentially could be further reduced by combining it with the above-mentioned methods (especially regarding more elaborate net architectures and regularization methods).

The CHiME\,3 corpus focuses on the application of speech recognition technologies in \textit{real-world} scenarios and sets a benchmark for comparing the value of  artificially generated data for training and testing purposes. Among the features tested, HTM yielded the lowest WER difference between the real and simulated test data, which indicates an improved generalization when combining DNNs with these features. Additionally, because both Aurora\,4 and CHiME\,3 tasks are based on WSJ and share some acoustic scenarios, conditions "D" (noisy, different microphone characteristics) from the former should be comparable with the (simu) condition from the latter. This is not the case for LTM and the whole Gabor filterbank which suggests the non-shared noises could particularly harm LTM and thus the whole Gabor filterbank as well.

Concerning the third CHiME challenge itself, every system in the top $10$ made several substantial changes to the baseline including augmentation of training data, speech enhancement, denoising, feature extraction, and improving or replacing the acoustic and language models. The top-ranked \citep{Yoshioka2015} included a pre-processing model based on spectral masking and beamforming; however, the deep learning architectures were trained on MFSC features. \citet{Vu2015} focused mainly on speech enhancement via non-negative matrix factorization and beamforming as well and also replaced the HMM decoder for a recurrent neural network. 

From the previous challenge, \citet{Moritz2013} indicated the modules developed in a hearing research environment were compatible and provided an incremental gain when combined; therefore the aforementioned systems could potentially benefit from the inclusion of auditory-based features as we show in this work.

Owing to the relatively low grammatical complexity in the first track of the CHiME\,2 Challenge, the contribution of the language model can be delimited, thus the WER depends more on the acoustic model trained on the features we want to compare. In this task, the error rates at the lowest SNR highlights the robustness of auditory-based features. Given that our error rates include complete utterances (not exclusively the letters and digits tokens), they are not directly comparable to the studies submitted to the challenge.

To get a better understanding of how relevant is the information provided to the DNN, the similarity analysis was conducted. It revealed HTM are able to better separate phoneme classes, potentially resulting in fewer confusions during classification. To support the results obtained from this analysis, we recurred to acoustic properties defined by \citet{Jakobson1956} (referred as distinctive features). 

Using these binary properties most confusions can be explained based on spectral properties of the phoneme classes; for instance, vowels and approximants confusions are not sufficiently covered in Jakobson/s distinctive features scheme. Even if there are articulatory and perceptual properties can be used to describe these classes, some confusion patterns observed on \fref{fig:CHiME2_SM}\,(c) such as: /IH/ \& /UW/, /IY/ \& /UW/ and /IH/ \& /R/ are unexpected as the phonemes involved are acoustically far apart from each other.

The similarity analysis on the Aurora\,4 corpus (shown in \fref{fig:Aurora4_SM}) exposed the discrimination capabilities of baseline MFSC and HTM. Because the distribution of vowels in the multidimensional space produce similar patterns for both features, we focused on consonant confusions to explain the improvements obtained with HTM, starting with the high-similarity pair /T/ \& /D/. Both phonemes share many acoustic properties, except the former has longer duration, reduced voice onset time, and higher total amount of energy with greater spread across the spectrum (typical characteristics of the disctintive feature known as \textit{tense}) and the latter, being \textit{voiced}, presents periodic low frequency excitation. 

The same distinction applies to the /P/ \& /B/ confusion, additionally, both phonemes have a energy in the lower frequencies (property denominated \textit{grave}) but only the /B/ presents energy on the closure phase, hence a steep transition is formed from the occlusion to the burst. This spectral change is enhanced by spectro-temporal features, thus we believe it is the main factor for the low number of confusions with HTM.  

In terms of acoustic properties, /K/ differs from /T/ in being \textit{compact} and \textit{grave}; the first property refers to the concentration of energy in a particular region of the spectrum. Once again, the smaller confusion from HTM could be due to an accurate detection of the spectral transition from burst to aspiration in frequencies below $2$\,kHz. Between /P/ \& /K/, the key distinction is the diffuse spectrum (opposed to \textit{compact}) of the former; therefore, this pair of phonemes is spectrally more similar than /T/ \& /K/ and thus has a higher similarity value.

The consistent confusion pattern observed among the plosives with /V/, /D/ and /DH/ presumably arises from their shared property \textit{voiced}. The periodic low frequency excitation, spectral tilt and burst frequency of stop consonants is severely deteriorated by additive noise even at medium signal-to-noise ratios. MFSC features encode the energy from the frequency bands so this effect is particularly harmful to these features. 

Finally, the similarity analysis on the CHiME\,2 allowed us to directly observe which confusions appear by decreasing the SNR. Note the analysis does not include the CHiME\,3 corpus because there is no clean data for the real recordings, which hinders generating high-quality phoneme labels with forced alignment as well as the calculation of the phoneme cluster statistics. In spite of sharing several phoneme confusions with Aurora\,4, there are some others to consider: /Z/ \& /S/ share almost all distinctive features except /Z/ is \textit{voiced}, this property can be adequately encoded by filters with a high spectral modulation. /B/ \& /G/ are voiced stops and their spectra bear some resemblance, although, in the case of /G/ it is \textit{compact}.

The phoneme /T/ is \textit{discontinuous} meaning there is an abrupt spectral transition, whereas /S/ is \textit{strident} presenting high energy noise dominated by high frequencies. While conceptually HTM could detect the abrupt transition of /T/, this property is mostly unnoticeable in isolation because it shows before the phoneme is pronounced. For the /IY/ \& /IH/ confusion, the former is \textit{tense}, which is a difficult distinctive feature to represent by either feature (also shown in \fref{fig:Aurora4_SM} for the /T/ \& /D/ confusion), as a longer temporal context might be needed and could account for appearing in all conditions. 

In \fref{fig:CHiME2_SM}\,(c), a consistent confusion pattern occurs for phoneme /F/ with respect to almost every phoneme from /N/ to /S/ (except phoneme /L/); this \textit{noisy} pattern is due to the negative the \textit{discontinuous} property, which is presumably why this pattern is not present in \fref{fig:CHiME2_SM}\,(d). The confusion pair /P/ \& /F/ is noticeable in all conditions despite being as well distinctively \textit{discontinuous}. We think a possible reason is that in some cases, for individual frames of $25$ms duration, the spectrum of the frication phase in /P/ resembles the one of a short /F/; for some other cases the sudden transient from the closure to the frication phase or from the frication to the aspiration phase of the /P/ is better detected by HTM, hence the lower similarity value.

\section{Conclusions}
\label{conclusions}

The present study assessed the contribution of Gabor features in combination with deep learning architectures. We found a subgroup of filters within the Gabor filterbank capable of reducing even further the word error rates in the three different recognition tasks (Aurora\,4, CHiME\,2 and CHiME\,3). The proposed HTM  outperformed the MFSC baseline. These features are capable of detecting quick spectro-temporal transitions within 40\,ms time windows and exhibited robustness against channel distortions, low signal-to-noise ratios and acoustically challenging \textit{real-life} scenarios; they also perform better on clean-conditions.

Because the gains presented in \tref{tab:WER} come from a relatively simple feature exchange, i.e. no additional speech enhancement, dereverberation or denoising techniques are applied, we assume it is straightforward to further improve the performance by combining one or more approaches including alternative deep learning approaches such as more elaborate net architectures and regularization methods.

The discriminability of MFSC and HTM was evaluated through the similarity analysis and explained in terms of distinctive spectral properties. The most relevant findings can be summarized as follows: Phonemes characterized as \textit{grave}, \textit{discontinuous} or \textit{compact} exhibit spectro-temporal transients, hence they are less likely to be confused by HTM features. \textit{voiced} consonants create consistent confusion patterns for MFSC features. \textit{tense} phonemes are equally hard to distinguish for both features. The confusions of obstruent consonants are more representative of the performance difference between HTM and MFSC features in the presence of additive noise and channel distortions. Overall HTM features produced a more separable distribution of phones.

Finally, in this study we show DNN-based speech recognizers trained with Gabor features, particularly the ones exclusively using high temporal modulation filters, yield lower error rates as these features enhance the discriminability between the target classes. 

% The information encoded by these filters can be used for human perceptual tasks or even signal reconstruction	

% Highlights:
%	DNN-based speech recognition greatly benefits from spectro-temporal Gabor features
%	Gabor filters with high temporal modulation encode the most relevant information
%	A measure of phoneme similarity is proposed to quantify class separability 
%	This metric is used to explain the improved results on phoneme level

% Long version
%   DNN-based speech recognizers trained with auditory-based Gabor features, particularly the ones exclusively using high temporal modulation filters yield lower word error rates
%   Similarity between phoneme classes let us quantify the discriminability of Mel-filterbank and Gabor features and learn why this approach works
%   The most relevant contribution from the Gabor filterbank is the high temporal modulation frequencies, this finding is supported by the similarity matrix patterns

\section*{Acknowledgment}

This work was funded by the DFG (Cluster of Excellence 1077/1 Hearing4All (http://hearing4all.eu), and the SFB/TRR 31 "The Active Auditory System" (http://www.sfb-trr31.uni-oldenburg.de/)) and by Google via a Google faculty award to Hynek Hermansky.

The authors thank Jon Barker for providing the clean labels for the CHiME\,2 Corpus and Mats Exter whose expert advice in phonetics and phonology was of great help for the discussion. We thank the reviewers for the time and effort invested in earlier drafts, whose helpful comments helped enrich and clarify this manuscript.

\vfill

\bibliographystyle{elsarticle-harv} 
\bibliography{refs}

\end{document}